\documentclass[10pt,twocolumn,letterpaper]{article}

\usepackage{cvpr}              % To produce the CAMERA-READY version
\usepackage{amsthm,amsmath,amssymb}
\usepackage{graphics}
\usepackage{tabularx}
\usepackage{multicol}
\usepackage{multirow}
\usepackage{diagbox}
\usepackage{makecell}
\usepackage{xcolor}
\usepackage{tabularx,utfsym,colortbl,arydshln}

\usepackage{color}
\usepackage{pifont}
\usepackage{enumitem}
\usepackage{stfloats}
\usepackage{caption}
\usepackage{adjustbox}
\usepackage{epstopdf}
\usepackage{wrapfig}
\usepackage[accsupp]{axessibility}

\newcommand{\methodname}{FedTSP}
\newcommand{\FPL}{FedPL}

\definecolor{cvprblue}{rgb}{0.21,0.49,0.74}
\usepackage[pagebackref,breaklinks,colorlinks,allcolors=cvprblue]{hyperref}

\def\paperID{5643} % *** Enter the Paper ID here
\def\confName{CVPR}
\def\confYear{2026}

\title{Enhancing Visual Representation with Textual Semantics: Textual Semantics-Powered Prototypes for Heterogeneous Federated Learning}

\author{
Xinghao Wu$^{1}$ \quad
Jianwei Niu$^{1,2}$ \quad
Xuefeng Liu$^{1,2}$\thanks{Corresponding author} \quad
Guogang Zhu$^{1}$ \quad \\
Jiayuan Zhang$^{1}$ \quad
Shaojie Tang$^{3}$ \quad
Wei Chen$^{1}$\\
$^{1}$State Key Laboratory of Virtual Reality Technology and Systems, \\ School of Computer Science and Engineering, Beihang University, Beijing, China\\
$^{2}$Zhongguancun Laboratory, Beijing, China\\
$^{3}$Center for AI Business Innovation, Department of Management Science and Systems, \\ School of Management, University at Buffalo, USA\\
{\tt\small \{wuxinghao, niujianwei, liu\_xuefeng, buaa\_zgg, zhangjiayuan, chenweibuaa\}@buaa.edu.cn} \\
{\tt\small shaojiet@buffalo.edu}
}

\begin{document}

\maketitle
\begin{abstract}
    Federated Prototype Learning (\FPL{}) has emerged as an effective strategy for handling data heterogeneity in Federated Learning (FL). In \FPL{}, clients collaboratively construct a set of global feature centers (prototypes), and let local features align with these prototypes to mitigate the effects of data heterogeneity. The performance of \FPL{} highly depends on the quality of prototypes. Existing methods assume that larger inter-class distances among prototypes yield better performance, and thus design different methods to increase these distances. However, we observe that while these methods increase prototype distances to enhance class discrimination, they inevitably disrupt essential semantic relationships among classes, which are crucial for model generalization. This raises an important question: how to construct prototypes that inherently preserve semantic relationships among classes?
Directly learning these relationships from limited and heterogeneous client data can be problematic in FL. Recently, the success of pre-trained language models (PLMs) demonstrates their ability to capture semantic relationships from vast textual corpora. Motivated by this, we propose FedTSP, a novel method that leverages PLMs to construct semantically enriched prototypes from the textual modality, enabling more effective collaboration in heterogeneous data settings. We first use a large language model (LLM) to generate fine-grained textual descriptions for each class, which are then processed by a PLM on the server to form textual prototypes. To address the modality gap between client image models and the PLM, we introduce trainable prompts, allowing prototypes to adapt better to client tasks.
Extensive experiments demonstrate that FedTSP mitigates data heterogeneity while significantly accelerating convergence. The code is available at \href{https://github.com/XinghaoWu/FedTSP}{https://github.com/XinghaoWu/FedTSP}.   
\end{abstract}
\vspace{-10pt}
\section{Introduction}
\label{sec:intro}

Federated Learning (FL) \citep{mcmahan2017communication} enables clients to collaboratively train models without exposing their raw data. A major challenge in FL is data heterogeneity, where data distributions across clients are non-independent and identically distributed (non-IID). Recently, Federated Prototype Learning (\FPL{}) has emerged as a promising approach to address data heterogeneity. In \FPL{}, all clients collaboratively construct a set of global feature centers (prototypes). These prototypes can serve as knowledge carriers \citep{tan2022fedproto,mu2023fedproc,tan2022federated,zhuAlignFed,zhang2024fedtgp,zhou2025fedsa} or function as feature anchors to align client representations \citep{zhu2022aligning,huang2023rethinking,xu2023personalized,ohfedbabu,li2023no,dai2023tackling,Qi_MM,Qi_AAAI} to reduce the impact of data heterogeneity.

It is evident that the performance of \FPL{} largely depends on the quality of the global prototypes. Early approaches typically aggregate client prototypes to obtain global prototypes \citep{tan2022fedproto, mu2023fedproc, 10286887, huang2023rethinking, wangtaming}. However, recent studies reveal that due to data heterogeneity, client prototypes often exhibit significant divergence, leading to suboptimal aggregated prototypes. Additionally, increasing the inter-class distance between prototypes is often associated with improved performance.
To this end, several methods propose manually designing or learning trainable prototypes on the server, aiming to maximize inter-class separation. For instance, AlignFed \citep{zhuAlignFed} and FedNH \citep{dai2023tackling} manually define server-side prototypes, ensuring that different class prototypes are uniformly distributed on a hypersphere. Meanwhile, FedTGP \citep{zhang2024fedtgp} and FedSA \citep{zhou2025fedsa} initialize trainable prototypes on the server, which are updated using client-side feature centers while enforcing larger inter-class distances.

\begin{figure*}[tb]
\setlength{\abovecaptionskip}{0.cm}
\setlength{\belowcaptionskip}{-0.cm}
        \centering  %图片全局居中

	\subcaptionbox{AlignFed}{\includegraphics[width=0.24\linewidth]{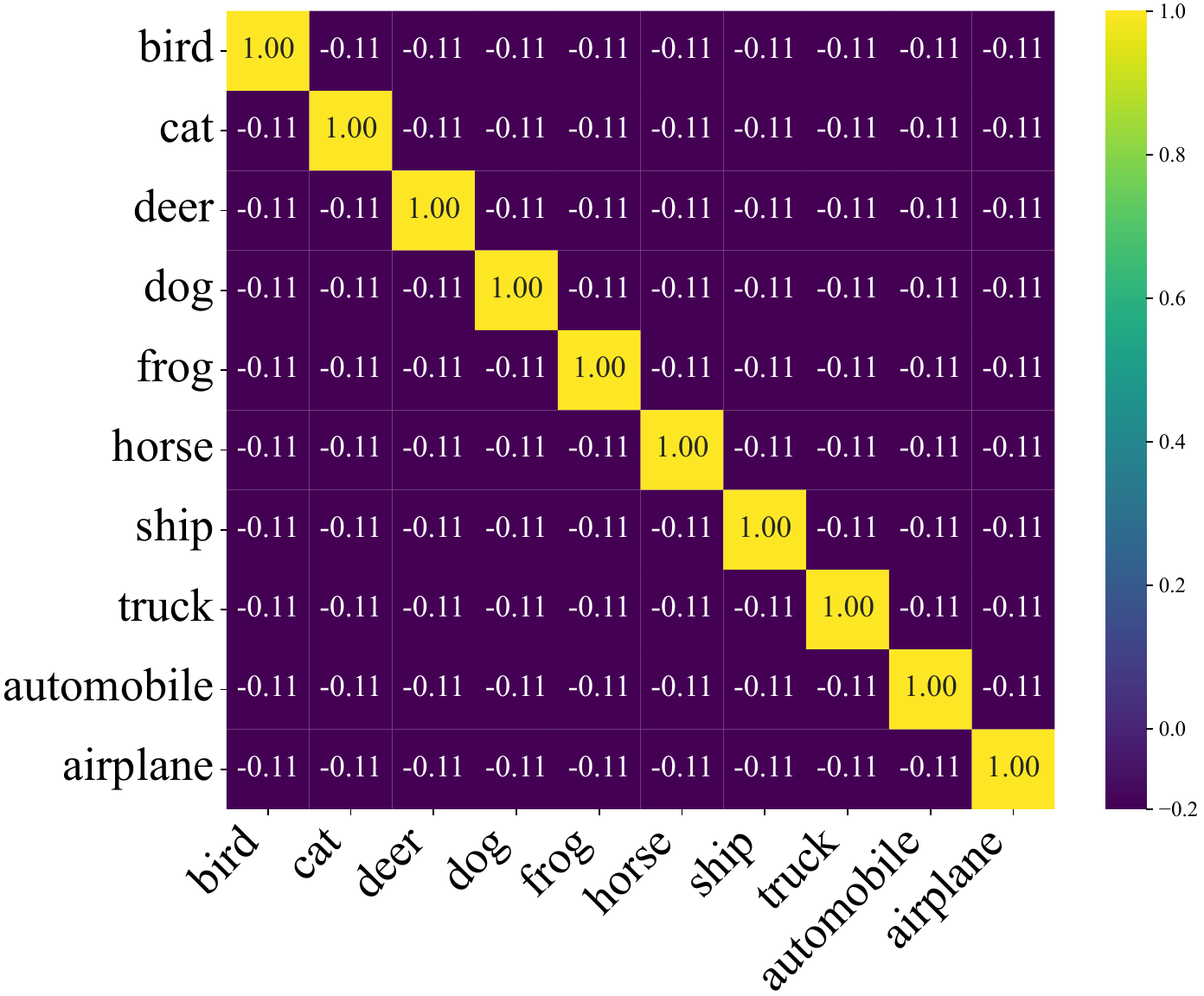}}
	\subcaptionbox{FedTGP}{
		\includegraphics[width=0.24\linewidth]{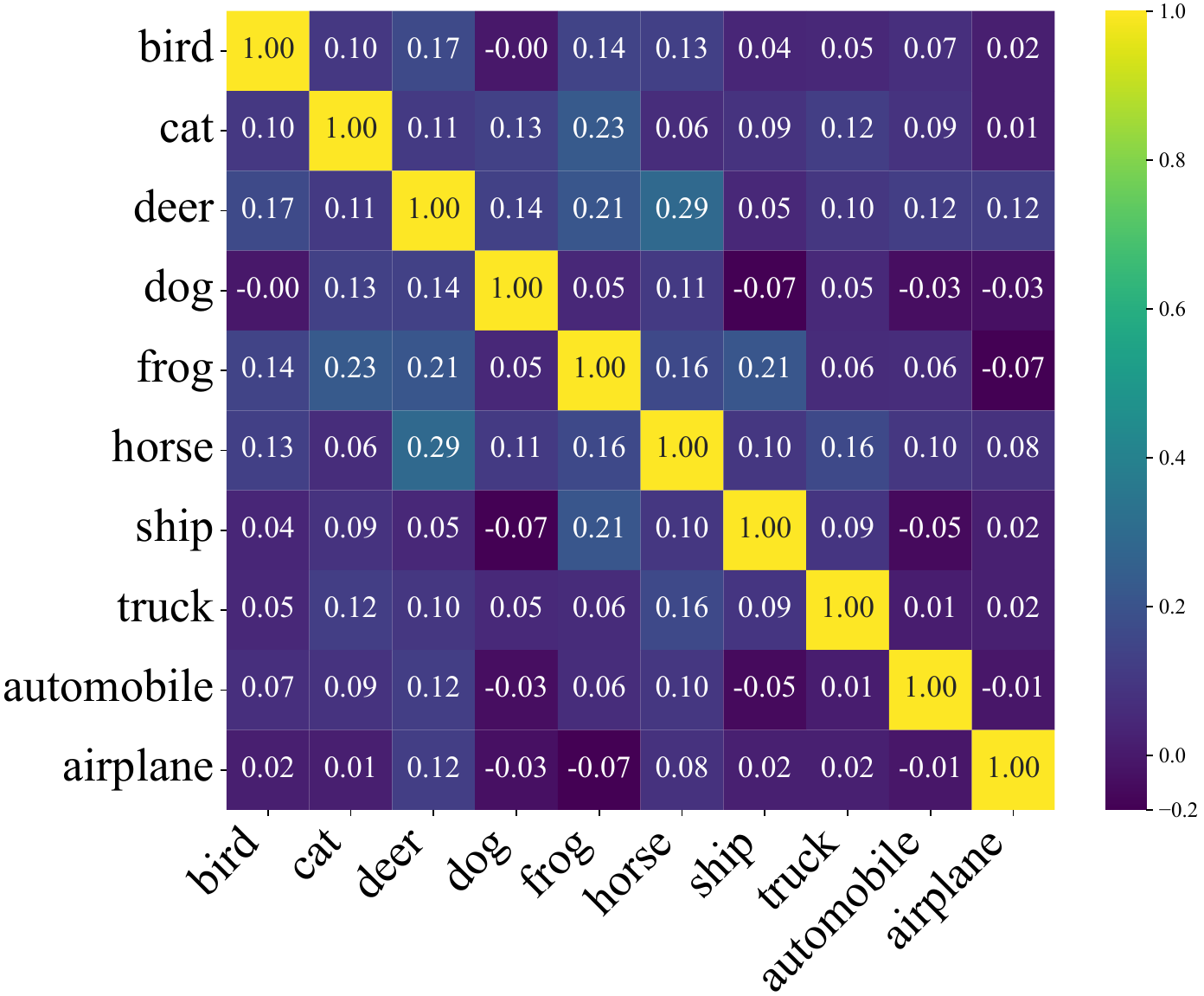}}
  \subcaptionbox{FedProto}{
		\includegraphics[width=0.24\linewidth]{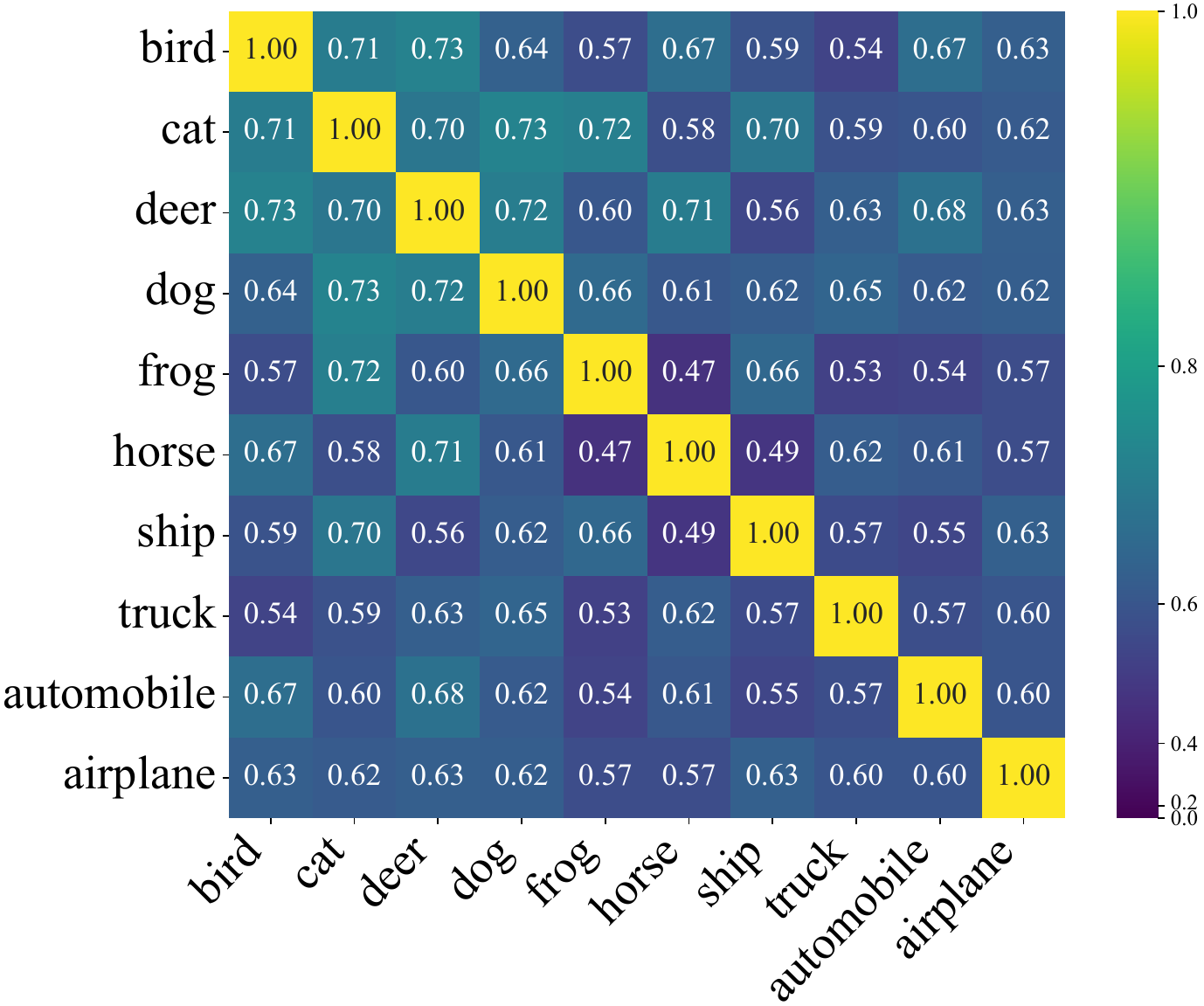}} 
    \subcaptionbox{Ours}{
		\includegraphics[width=0.24\linewidth]{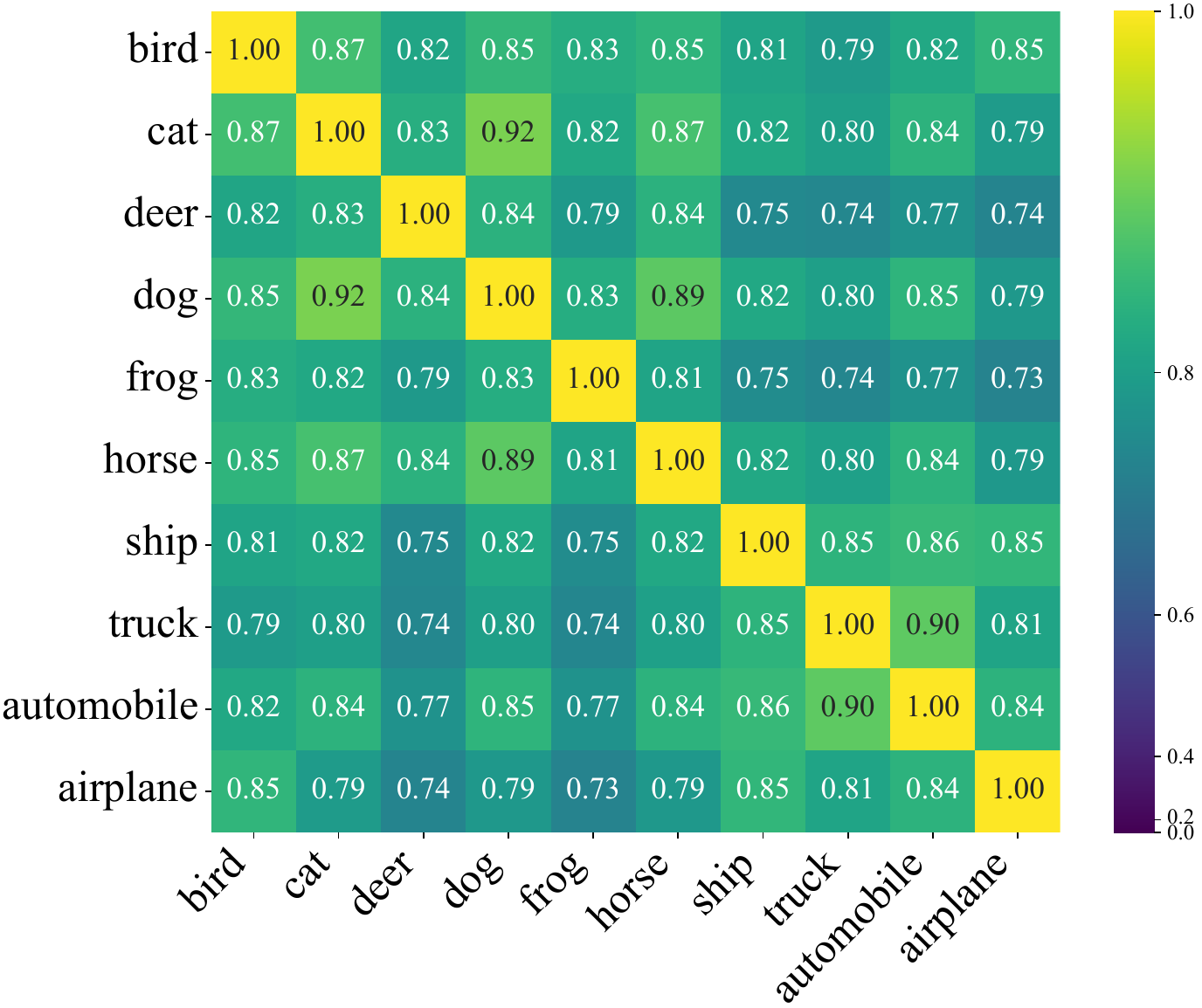}} \\
        \subcaptionbox{Semantic Alignment ($\uparrow = \text{better}$)}{
		\includegraphics[width=0.49\linewidth]{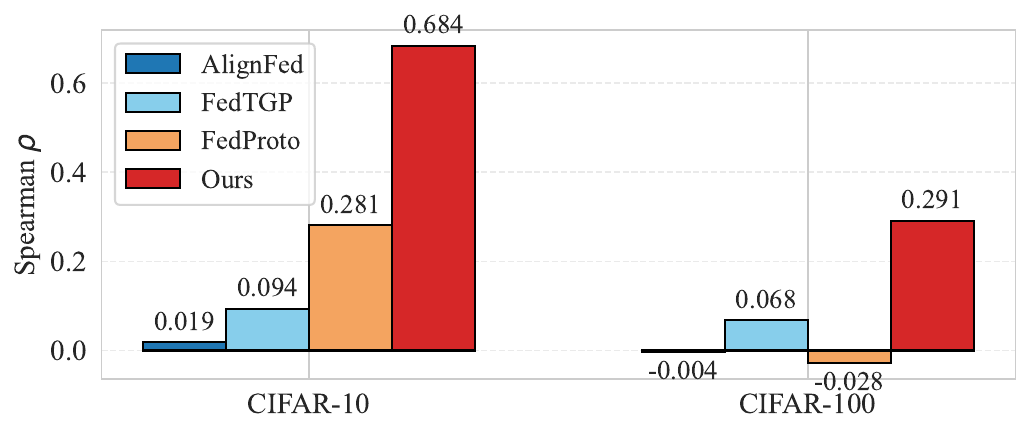}}
        \subcaptionbox{Semantic Gap ($\uparrow = \text{better}$)}{
		\includegraphics[width=0.49\linewidth]{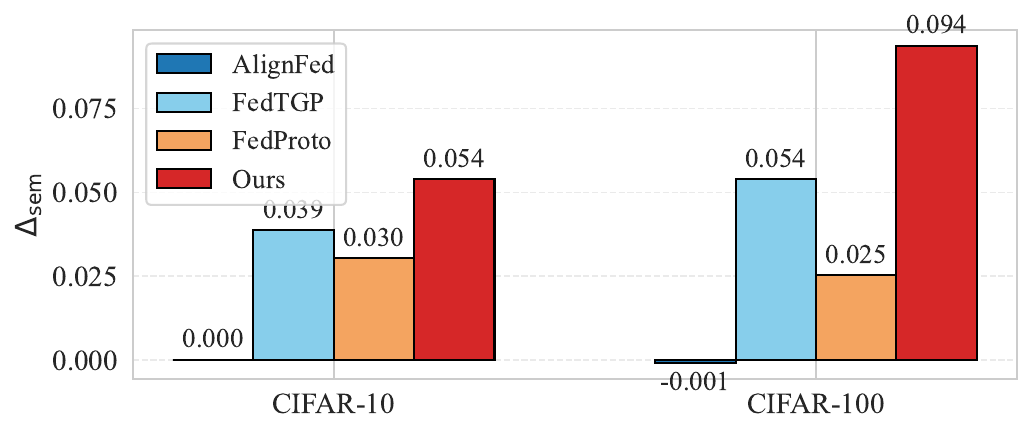}}
	\caption{\textbf{Semantic structure of learned prototypes.}
(a–d) show the cosine similarity between prototypes of different classes on CIFAR-10. (e–f) provide quantitative evaluations of the semantic structure. The metrics in (e–f) are computed from the similarities shown in (a–d) and extended to CIFAR-100 for broader validation. (e) shows the Spearman correlation $\rho$ between model-learned similarity rankings and WordNet-derived semantic rankings. (f) reports the semantic gap $\Delta_{\text{sem}}$, defined as the difference between the average similarity within semantic groups (e.g., animals vs. animals) and across different groups (e.g., animals vs. vehicles).  Our method consistently yields stronger alignment with human-perceived semantics and larger semantic gaps across both datasets.}
	\label{fig:motivation}
 \vspace{-0.15in}
\end{figure*}

However, we observe that existing prototype‐based FL methods often fail to preserve class-level semantic structure, which is crucial for model generalization. Intuitively, prototypes of semantically related classes (\eg, horse and dog) should be closer than those of unrelated classes (\eg, horse and truck). Such semantic consistency enables models to exploit feature commonality across related classes while distinguishing unrelated ones. Figs.~\ref{fig:motivation}(a–c) visualize prototype–prototype cosine similarity on CIFAR-10 for three representative methods. AlignFed and FedTGP yield either uniformly distributed or overly dispersed prototypes, whereas FedProto learns some structure from client data but still shows visible mismatches due to data heterogeneity.

To quantify these observations, we compute two metrics from the similarities in Figs.~\ref{fig:motivation}(a–d): 
(i) the \emph{semantic alignment score}, given by Spearman correlation $\rho$ between model-learned similarity rankings and semantic rankings derived from WordNet~\citep{miller1995wordnet}, a widely-used lexical database that encodes human-defined semantic relationships among words;
and (ii) the \emph{semantic gap} $\Delta_{\text{sem}}$, i.e., the difference between average similarity within semantic groups (animals vs. animals, vehicles vs. vehicles) and across groups (animals vs. vehicles).  As Figs.~\ref{fig:motivation}(e–f) show, the alignment and gap of prior methods remain limited on CIFAR-10 and degrade further on CIFAR-100.

Recent advancements in pre-trained language models (PLMs) such as BERT \citep{Bert} and large language models (LLMs) like ChatGPT \citep{openai2023chatgpt} suggest that textual embeddings encode rich, human-level semantics. This motivates us to ask: \textbf{Can we inject textual semantic knowledge into FL prototypes so that they retain class relationships even under heterogeneous data?}  

We answer this question with \methodname{}, a new \FPL{} framework that constructs class prototypes from the textual modality. \methodname{} incorporates semantic knowledge from text models into FL by deploying a PLM on the server. This PLM generates text prototypes based on textual descriptions (prompts) for each class. To enrich the semantic information encoded in these prototypes, \methodname{} leverages an LLM to generate detailed prompts, rather than relying on handcrafted ones. As evidenced by the higher Spearman $\rho$ and larger $\Delta_{\text{sem}}$ in Figs.~\ref{fig:motivation}(e–f), the resulting text prototypes effectively preserve semantic relationships significantly better than prior approaches. Furthermore, since the PLM may have never been exposed to image data during pretraining, a modality gap exists between the PLM and client-side image models. To bridge this gap, \methodname{} introduces a set of trainable prompts within the PLM, enabling the generated prototypes to better adapt to client tasks. By aligning local features with these prototypes, client models can effectively learn textual semantic knowledge.
% inter-class semantic relationships.

Our key contributions can be summarized as follows:
\begin{itemize}
\item We identify that existing FedPL methods disrupt semantic relationships among similar classes and propose leveraging these relationships to enhance model performance.
\item We propose \methodname{}, a novel method that introduces semantic knowledge from PLMs and LLMs to enhance FL training. To the best of our knowledge, this is the first work to integrate text-derived semantics into \FPL{}.
\item We conduct extensive experiments on multiple datasets across diverse heterogeneity settings. The results demonstrate that \methodname{} not only outperforms existing methods but also significantly accelerates convergence.
\end{itemize}
\vspace{-10pt}
\section{Related Work}\label{sec:related}
\textbf{Federated Prototype Learning.}
In \FPL{}, clients collaboratively construct global prototypes to facilitate feature alignment and mitigate data heterogeneity. Early approaches such as FedProto \citep{tan2022fedproto}
, FedProc \citep{mu2023fedproc}, FedFA \citep{10286887}, and FedPAC \citep{xu2023personalized}, aggregate client-specific prototypes via federated averaging. 
%However, due to data heterogeneity, direct averaging may lead to biased global prototypes. To address this, cluster-based methods like FPL \citep{huang2023rethinking} and FedPLVM \citep{wangtaming} first cluster similar client prototypes before aggregation to enhance robustness.
Recent studies highlight that data heterogeneity causes large variations in client prototypes, making naive aggregation suboptimal. Moreover, increasing inter-class separation has been shown to improve performance. To this end, methods like AlignFed \citep{zhuAlignFed} and FedNH \citep{dai2023tackling} manually define uniformly distributed prototypes on a hypersphere, while FedTGP \citep{zhang2024fedtgp} and FedSA \citep{zhou2025fedsa} introduce trainable server-side prototypes, updated using client feature centers. Additionally, FedBABU \citep{ohfedbabu} and FedETF \citep{li2023no}, randomly initialize or employ ETF-based initialization for a fixed global classifier. Since classifier proxies inherently serve as a set of class prototypes, these methods can also be regarded as a form of prototype learning.
%Despite these advancements, existing methods overemphasize inter-class separation while neglecting semantic relationshipsfe, which are crucial for generalization. Thereby, we propose incorporating PLMs and LLMs to construct semantically enriched prototypes, providing better guidance for client learning.

\noindent\textbf{Heterogeneous Federated Learning} (HtFL) \cite{HtFLlib} enables client collaboration under both data and model heterogeneity, where a key challenge is extracting global knowledge without direct model aggregation. Existing HtFL methods mainly fall into three categories: parameter-sharing \citep{liang2020think,zhu2021data,yi2023fedgh}, knowledge distillation \citep{shen2020federated,wu2022communication,yi2024federated}, and prototype-based \citep{tan2022fedproto,zhang2024fedtgp,zhang2024upload,zhou2025fedsa} approaches. Among them, prototype-based methods stand out due to their lower computational and communication costs, as clients only need to share class-wise feature centers.

\noindent\textbf{CLIP-based Federated Learning} primarily focuses on improving CLIP's performance on client-specific downstream tasks. These approaches typically adapt CLIP via prompt tuning \citep{guo2023promptfl,guo2023pfedprompt,li2024global,qiu2024federated,Qi_NIPS} or contrastive alignment \citep{yanimportance,Shi_AAAI}, and rely on the fine-tuned CLIP model itself for inference. In contrast, our work adopts a fundamentally different perspective: rather than strengthening CLIP, we aim to transfer the semantic structure encoded in pretrained language models (PLMs or LLMs) into lightweight, client-specific image models. In our setting, the client model performs the downstream task. This shift makes our method agnostic to CLIP, scalable to heterogeneous and resource-constrained clients, and extensible to non-vision-language PLMs such as BERT. While CLIP facilitates semantic transfer through its vision-language alignment, it is not required for our framework.

%This work focuses on the HtFL scenario, where both data and model heterogeneity coexist, further complicating the construction of robust prototypes.
\vspace{-6pt}
\section{Methodology}\label{sec:method}
In this section, we elaborate on the proposed \methodname{}. We mainly formulate it under the model heterogeneous scenario, which is a more challenging setting for constructing globally consistent prototypes that preserve semantic relationships. The extension to homogeneous settings is provided in Section~\ref{sec:extend to homo}.
%In this section, we provide a detailed explanation of \methodname{}. We \textbf{formulate the method under the HtFL setting}, which poses the greatest challenge due to the presence of \textbf{both data heterogeneity and model heterogeneity} across clients. This heterogeneity further exacerbates the difficulty of constructing robust prototypes. In Section~\ref{sec:extend to homo}, we demonstrate that \methodname{} can be easily extended to both the General FL (GFL) and Personalized FL (PFL) settings.
% \vspace{-10pt}
\subsection{Preliminaries}
\textbf{Heterogeneous Federated Learning.}
In HtFL, $N$ clients collaborate under the coordination of a central server to train their respective \text{heterogeneous personalized models} $\{w_i\}_{i=1}^N$. Each client aims to achieve strong performance on its own data distribution $\mathbb{D}_i$, where $\mathbb{D}_i \neq \mathbb{D}_j$ for any $i \neq j$. Following FedProto and FedTGP, we decompose $w_i$ into a feature extractor $f_i: \mathcal{X}^{\mathbf{I}} \rightarrow \mathbb{R}^d$, parameterized by $\theta_i$, and a classifier $h_i: \mathbb{R}^d \rightarrow \mathbb{R}^C$, parameterized by $\phi_i$. Here, $\mathcal{X}^{\mathbf{I}}$ denotes the space of raw image data, $d$ is the feature dimension, and $C$ is the number of classes. The optimization objective of HtFL is formulated as follows:
\begin{equation}
\min_{\{\theta_i, \phi_i\}_{i=1}^N} \quad \frac{1}{N}\sum_{i=1}^{N} \mathcal{L}_D\bigl(\theta_i, \phi_i; \mathbb{D}_i \bigr),
\end{equation}
where $\mathcal{L}_D$ denotes an empirical loss function. 
% (\eg, cross-entropy).

% Due to the limited number of samples in each client’s training set $\mathcal{D}^{\text{train}}_i$, directly optimizing $\mathcal{L}_D$ locally can lead to overfitting, thereby hindering generalization to the true distribution $\mathbb{D}_i$. Therefore, effectively enabling client collaboration to enhance generalization in scenarios with both model and data heterogeneity is a key challenge in HtFL.

\noindent\textbf{Federated Prototype Learning.}
In \FPL{}, all clients collaborate by sharing a global prototype $\mathcal{P}=\{\mathcal{P}^c\}_{c=1}^{C}$, aiming to guide feature extraction through the jointly constructed feature structure. During local training, clients learn the feature structure by aligning their local features with $\mathcal{P}$. The training objective can be formulated as:
\begin{equation}\label{eq:FPL obj}
    \mathcal{L}_i = \mathcal{L}_D(\theta_i, \phi_i; x, y) + \lambda \mathcal{R}(f_i(x; \theta_i), \mathcal{P}^y), \text{where } x, y \sim \mathcal{D}_i^{\text{train}}.
\end{equation}
$\mathcal{R}$ denotes the alignment loss used to constrain the alignment between client features and $\mathcal{P}$. $\mathcal{D}_i^{\text{train}}$ represents local training data. $\lambda$ is a hyperparameter used to balance the influence of $\mathcal{L}_D$ and $\mathcal{R}$.

% Common choices for $\mathcal{R}$ in current research include Mean Squared Error (MSE) loss \cite{} and contrastive learning loss \cite{}. $\lambda$ is a hyperparameter used to balance the influence of $\mathcal{L}_D$ and $\mathcal{R}$.

The quality and characteristics of $\mathcal{P}$ directly influence the effectiveness of client collaboration and the overall model performance. In this work, we focus on constructing $\mathcal{P}$ with richer semantic relationships to capture inter-class similarities and enhance the collaborative learning process.

% Current methods for constructing $\mathcal{P}$ primarily focus on increasing the pair-wise distance between classes, often neglecting the semantic relationships between different classes. 

\noindent\textbf{Text Features Extracted by a PLM.}
Let $\psi$ denote the feature extractor of a PLM. It typically consists of two main functional modules, expressed as $\psi = g_1 \circ g_0$. Here, $g_0$ represents the tokenizer and embedding layers, while $g_1$ serves as the backbone of the text encoder (commonly a Transformer \cite{vaswani2017attention} architecture). Given a text sequence $x^{\mathbf{T}}$ (the prompt) of length $n$, $g_0$ first maps it into an embedding matrix $\mathbf{E}=g_0(x^{\mathbf{T}}) \in \mathbb{R}^{n \times d^{\prime}}$ (\ie, a sequence of $n$ embedding vectors, each of dimension $d^{\prime}$). These embeddings are then fed into the backbone network to obtain a feature $\mathbf{H}=g_1 \circ g_0(x^{\mathbf{T}}) \in \mathbb{R}^{n \times d}$ (\ie, a sequence of $n$ hidden states, each of dimension $d$).

For many downstream tasks, we often need a single vector feature $\mathbf{z} \in \mathbb{R}^{d}$ derived from $\mathbf{H}$. For example, $\mathbf{z}=\mathbf{H}[0:]$ in BERT \cite{Bert} and $\mathbf{z}=\mathbf{H}[\text{EOS index},:]$ in CLIP \cite{CLIP}. For simplicity of expression, we incorporate this selection process into  $g_1$  in the subsequent text, resulting in  $\mathbf{H} = g_1 \circ g_0(x^{\mathbf{T}}) \in \mathbb{R}^d$.

% \vspace{-15pt}
\subsection{Overview of \methodname{}}
\begin{figure*}[tb]
% \vspace{-10pt}
	\setlength{\abovecaptionskip}{0.cm}
\setlength{\belowcaptionskip}{-0.cm}
    \centering
    \includegraphics[width=0.9\linewidth]{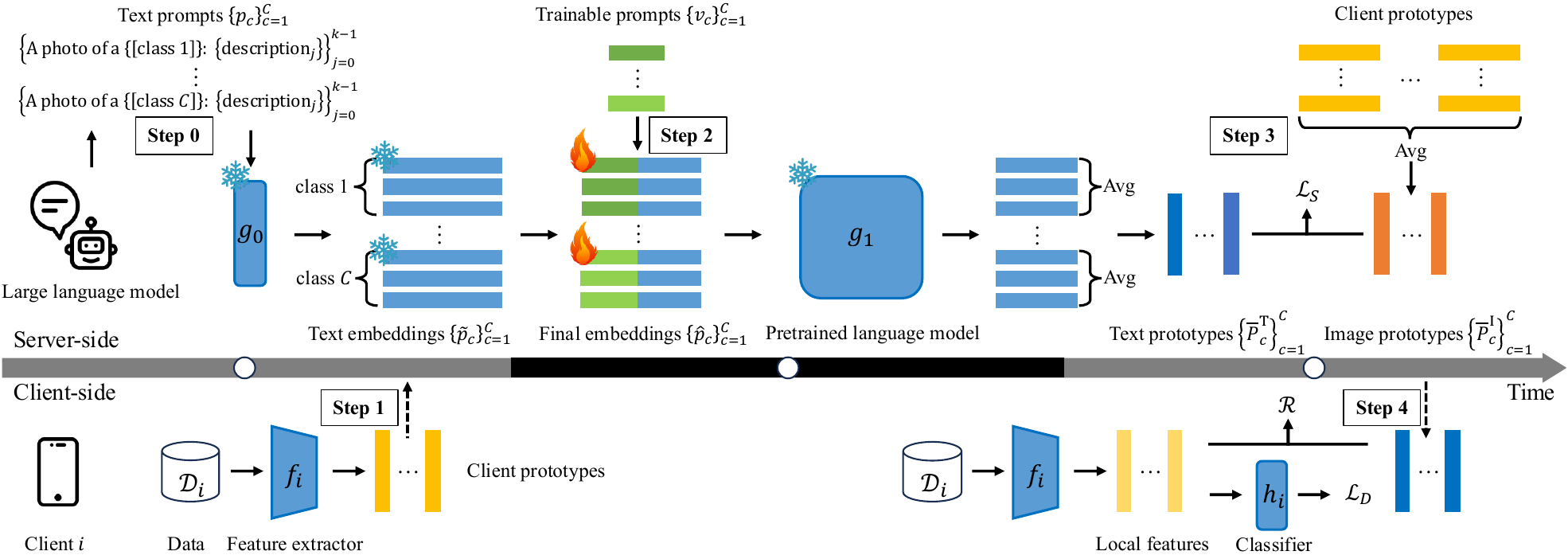}
    \caption{Overview of \methodname{}.}
    \label{fig:overview}
    \vspace{-20pt}
\end{figure*}

Before delving into the details, we present the overall workflow of \methodname{}, as illustrated in Fig.~\ref{fig:overview}.

\textbf{Step 0}: The server queries an LLM API to generate class descriptions and construct text prompts, which are then encoded into text embeddings by the pre-trained PLM.

\textbf{Step 1}: Each client computes image prototypes and uploads them to the server.

\textbf{Step 2}: The server inserts a set of trainable prompts into the text embeddings to obtain the final embedding. The PLM then extracts features to generate text prototypes.

\textbf{Step 3}: The server aggregates client prototypes to obtain the image prototypes and updates the trainable prompts to align the text prototypes with the image prototypes.

\textbf{Step 4}: The server distributes the updated text prototypes to clients, which then optimize their local objectives while aligning local features with the text prototypes.

Steps 1 to 4 are repeated until convergence.

% \subsection{Problem Formulation}

% \subsection{Text Prototype Generation}
\subsection{Injecting Textual Semantics into Prototypes}
%Unlike existing methods that typically build or train prototypes directly from client image data, 

\methodname{} leverages a PLM to generate text prototypes for each class to harness the rich semantic relationships presented in the text modality. These prototypes can more effectively guide feature extraction on the client side, resulting in a more informative feature structure.

% \noindent\textbf{LLM-based Prompt Generation.}
\noindent\textbf{Generating Multi-View Prompts with LLMs.}
%A key ingredient of \methodname{} is the introduction of semantic information from the text modality. Consequently, one central challenge is designing effective prompts $p_c$ for each class $c$ that encapsulate richer semantic details and capture fine-grained distinctions between classes. 
Existing works typically adopt CLIP’s zero-shot prompt template, \eg, “A photo of a \{CLASS\}”, where the \{CLASS\} token is filled with the class name. However, this design offers only minimal descriptive detail: prompts for different classes differ solely in the class name, providing limited semantic context and restricting the model’s ability to capture more nuanced characteristics. In addition, relying exclusively on a class name often introduces ambiguity. For instance, the term “apple’’ may denote either a fruit or a technology company, which can hinder the text encoder from accurately learning class-level semantic similarities.

To address these limitations, we propose querying an LLM API on the server side to generate more fine-grained descriptions for each class. Rather than relying on handcrafted prompts, \methodname{} adopts the template \text{“A photo of \{CLASS\}: \{text description\}”}, where {text description} is generated automatically by an LLM (\eg, ChatGPT). For each class, \methodname{} produces $k$ descriptions covering different aspects of the class. This strategy helps the text encoder capture finer-grained semantic distinctions. Formally, each class $c$ is associated with a prompt set:
\begin{equation}
p_c = \bigl\{\,p_{c,j}\,\big|\,\text{``A photo of \{CLASS\}:} \{\text{description}_j\}\text{''} ,{j \in [k]}\bigr\}.
\end{equation}
By incorporating these multiple perspectives, \methodname{} enriches the semantic context, guiding the image model to learn more discriminative features for each class. In our implementation, we set  $k = 3$  as a fixed value.

% \noindent\textbf{Trainable Prompt for Downstream Task Alignment.}
\noindent\textbf{Learnable Prompts for Modality Alignment.}
Although pretrained text encoders typically embed rich semantic knowledge, most of them have not been trained on image data during pretraining. This can lead to a modality gap between text features and image features. To address this issue, we introduce trainable prompts \cite{zhou2022conditional} within the text prompts, which are updated during training to align with the clients’ visual tasks.

Formally, for a text prompt $p_c$ corresponding to class $c$, let $\tilde{p}_c = g_0(p_c) \in \mathbb{R}^{k \times n \times d^{\prime}}$ denote its embedded feature. We introduce a set of trainable embedding vectors $v_c \in \mathbb{R}^{m \times d^{\prime}}$ and replace the first $m$ segments of $\tilde{p}_c$:
\begin{equation}
\hat{p}_c =\bigl\{
\bigl[\;\underbrace{v_{c,0}, v_{c,1}, \ldots, v_{c,m-1}}_{\text{trainable prompts}}, \;\tilde{p}_{c,j}\bigl[m:\bigr]\bigr]\bigr \}_{j=0}^{k-1},
\end{equation}
where $\hat{p}_c$ is the final embedding sequence for class $c$. Notably, the $k$ text prompts for a single class $p_c$ share a common set of trainable prompts $v_c$. The text encoder processes $\hat{p}_c$ via $g_1(\hat{p}_c)$ to generate class-specific features $P_{c}^{\mathbf{T}} \in \mathbb{R}^{k \times d}$. The text prototype $\overline{P}_c^{\mathbf{T}} \in \mathbb{R}^d$ is then obtained by averaging the features derived from the $k$ prompts for each class $c$.

At the end of each training round, each client computes its local image prototypes by
\begin{equation}
    P_{i,c}^{\mathbf{I}} = \frac{1}{|\mathcal{D}^{\text{train}}_{i,c}|} \sum_{(x,y)\in \mathcal{D}^{\text{train}}_{i,c}} f_i(x;\theta_i), c\in[C]
\end{equation}
and sends them to the server. The server aggregates these local image features to derive global image prototypes:
\begin{equation}
    \overline{P}_c^{\mathbf{I}} = \frac{1}{|\mathcal{M}_c|} \sum_{i \in \mathcal{M}_c} \frac{|D_{i,c}^{\text{train}}|}{M_c} P_{i,c}^{\mathbf{I}},
\end{equation}
where $\mathcal{M}_c$ denotes the set of clients that have class $c$, and $M_c$ represents the total number of samples of class $c$ across all clients. These global prototypes $\overline{P}_c^{\mathbf{I}}$ are subsequently used to update the trainable prompts by
\begin{align}
    \min_{\{v_c\}_{c=0}^{C-1}} &\mathcal{L}_{S}(\{v_c\}_{c=0}^{C-1},\psi;\{p_c\}_{c=0}^{C-1}) \\ \nonumber
    &=-\frac{1}{C}\sum_{c=0}^{C-1} \log \frac{\exp (\cos(\overline{P}_c^{\mathbf{T}}, \overline{P}_c^{\mathbf{I}})/\tau)}{\sum_{j=0}^{C-1} \exp (\cos(\overline{P}_c^{\mathbf{T}}, \overline{P}_j^{\mathbf{I}})/\tau)}
\end{align}
for $E_s$ epochs.
This design aligns text prototypes with image prototypes on the server-side, allowing textual features to capture task-specific visual cues and bridging the modality gap between the PLM and client image models. 
% Ultimately, this mechanism enhances model performance in HtFL scenarios.

% \subsection{Local Training}
\subsection{Transferring Prototype Semantics to Clients}
As defined in Eq.~\eqref{eq:FPL obj}, on the client side, each client’s objective is to train the model $w_i$ on its local dataset while aligning the locally extracted features with the global prototypes. 
In this work, we primarily focus on classification tasks, \ie, $\mathcal{L}_D$ is defined as the cross-entropy loss.
% \begin{align}
% \mathcal{L}_D(\theta_i,\phi_i; x, y) = & - \sum_{c=1}^{C} \mathbb{I}(y=c) y_{c}\log\bigl(\sigma(h_i \circ f_i(x)) \bigr), 
% % &\text{where } x, y \sim D^{\text{train}}_i \nonumber
% \end{align}
% where $\mathbb{I}$ is indicator function, $\sigma$ denotes the softmax function.

For feature alignment, as shown in Fig.~\ref{fig:motivation}(d), PLM effectively captures the relative similarity between classes (\eg, animals are more similar to each other than to vehicles). However, it tends to exhibit high baseline similarity across all classes, with even the least similar classes still having high similarity scores (\eg, 0.73). If L2-based feature alignment \cite{tan2022fedproto,zhang2024fedtgp} is applied, it optimizes for absolute similarity, which can mislead the model into treating unrelated classes as similar, ultimately degrading semantic precision.

To address this, \textit{we focus on relative similarity among classes} and propose an optimization method based on contrastive learning loss \cite{chen2020simple}. Contrastive learning minimizes the distance between positive pairs (semantically related classes) and maximizes the distance between negative pairs (semantically unrelated classes). The temperature parameter $\tau$ scales the similarity values, enabling the model to prioritize the relative ranking of similarities rather than absolute values. Specifically, the alignment term $\mathcal{R}$ in \methodname{} is defined as:
\begin{equation}
\mathcal{R}(\theta_i, x, y) = - \log \frac{\exp\bigl(\cos(f_i(x), \mathcal{P}^{\mathbf{T}}_y) / \tau\bigr)}{\sum_{c=0}^{C-1} \exp\bigl(\cos(f_i(x), \mathcal{P}^{\mathbf{T}}_c) / \tau\bigr)}.
\end{equation}

By aligning local features with the text prototypes, the local model inherits both the fine-grained semantic information and the inter-class semantic relationships learned from large text corpora. The pseudo-code of \methodname{} is summarized in the Appendix~A.
% Appendix~\ref{sec:algorithm}.

\subsection{Privacy-Preserving Extension of \methodname{}}\label{sec:privacy protection}

In the original design of FedTSP, textual descriptions for each class are generated on the server. This implicitly assumes that the server has access to all class names of the clients’ tasks. Although many prior works in FL make this assumption \cite{zhang2023navigating,zhu2024stealing,wang2025federated} and do not regard class names as sensitive information, this may still pose privacy risks in domains with strict confidentiality requirements.

To address this issue, we introduce a privacy-preserving variant of FedTSP by making a minimal architectural modification and incorporating differential privacy (DP). Since the generation of class descriptions is performed only once before training begins (\ie, Step 0 in Fig. 2), this step can be safely relocated to one selected client. Specifically, the client queries an LLM API to generate class descriptions, encodes them into text embeddings $\mathbf{E} \in \mathbb{R}^{n \times d'}$ using the PLM’s embedding layer $g_0$, and uploads the resulting embeddings to the server.

To prevent potential reconstruction of class names from these embeddings, we adopt a single-shot local $(\epsilon, \delta)$-DP mechanism \cite{wahdany2025differentially,chen2023federated}. The embedding sequence $\mathbf{E}$ is flattened into $\text{Vec}(\mathbf{E}) \in \mathbb{R}^{nd'}$, clipped to an L2 radius $C$, and perturbed with isotropic Gaussian noise:
\begin{equation}
    Y=\text{Clip}_C (\text{vec}(\mathbf{E})) + Z, Z \sim \mathcal{N}(0, \sigma^2 I _{nd'}).
\end{equation}
Here, $\text{Clip}_C(v)=v \cdot \min \{1, C / ||v||_2 \}$. Since $||\tilde{x}|| _2 \le C$ for any input, the L2 sensitivity satisfies
\begin{equation}
    \Delta = \sup _{X \sim X'} ||\tilde{x} - \tilde{x}'|| _2 \le 2C.
\end{equation}
By the Gaussian noise mechanism, choosing
\begin{equation}
    \sigma \ge \frac{\Delta \sqrt{2 \ln(1.25 / \delta)}}{\epsilon} = \frac{2C \sqrt{2 \ln(1.25 / \delta)}}{\epsilon}
\end{equation}
ensures single-shot local $(\epsilon, \delta)$-DP for the released embedding.
This modification protects class identity privacy while keeping the communication cost and computational burden minimal. In Section~\ref{sec:privacy expe}, we further evaluate the impact of this privacy-preserving mechanism on model performance.

% During local training, each client simultaneously updates both the feature extractor and the classifier via gradient descent using Eq.\eqref{}. Note that $\mathcal{L}_D$ updates both the feature extractor and the classifier, while $\mathcal{R}$ only updates the feature extractor.

\subsection{Extension to Model-Homogeneous Settings}\label{sec:extend to homo}
Although the proposed framework is primarily developed for model-heterogeneous FL, it can be readily applied to model-homogeneous cases such as General FL (GFL) and Personalized FL (PFL). 
In GFL~\cite{mcmahan2017communication,li2020federated,li2021model,zhang2024kdd,zhang2026aggregated,zhu2024ijcai}, since all clients’ features are aligned with shared text prototypes, local models exhibit reduced divergence and can be aggregated using FedAvg~\cite{mcmahan2017communication} to form a global model. 
For PFL~\cite{wu2023bold,FedDecomp,DiversiFed,FedPFT,zhang2025causality,zhu2024mm,zhu2024tnnls}, once a global feature extractor is learned, fine-tuning the classifier on each client yields personalized models~\cite{ohfedbabu,li2023no}. 
As demonstrated in Section~\ref{sec:GFL and PFL expe}, \methodname{} consistently outperforms state-of-the-art methods in both settings.

% NeurIPS version
% Although the previous sections primarily focus on scenarios with model heterogeneity, our approach can naturally be extended to model-homogeneous settings, GFL and PFL.

% GFL \cite{mcmahan2017communication,li2020federated,li2021model} trains a single global model for all clients. In \methodname{}, since all clients’ features are aligned with global text prototypes, the divergence among local models is reduced. We can therefore obtain a global model by simply aggregating the local models using FedAvg \cite{mcmahan2017communication}.

% PFL \cite{li2021ditto,wu2023bold,Zhang_2023_ICCV,zhang2023fedala,zhang2024eliminating,zhang2024enabling} trains a personalized model for each client. Recent PFL works have shown that once a global feature extractor is well-trained, fine-tuning only the classifier can yield strong personalized performance \cite{ohfedbabu,li2023no}. Therefore, we can directly fine-tune the classifier after obtaining the global model to achieve personalization. In Section~\ref{sec:GFL and PFL expe}, we demonstrate that \methodname{} outperforms state-of-the-art (SOTA) methods in both GFL and PFL settings.

% \vspace{-5pt}
\section{Experiments}\label{sec:experiment}
\subsection{Experimental Setup}
\textbf{Datasets}. We evaluate the effectiveness of \methodname{} on three benchmark image classification datasets: CIFAR-10 \cite{krizhevsky2010cifar}, CIFAR-100 \cite{krizhevsky2009learning}, and Tiny ImageNet \cite{le2015tiny}.

\noindent\textbf{Data Heterogeneity}. We adopt the Dirichlet non-IID setup \cite{wu2022pfedgf,wu2023bold,shi2023prior,FedDecomp,DiversiFed,FedPFT} in experiments, where each client’s data is sampled from a Dirichlet distribution, Dir($\alpha$). Here, $\alpha$ is a hyperparameter that controls the level of data heterogeneity, where a smaller $\alpha$ indicates a higher degree of heterogeneity. 
%As $\alpha$ increases, the imbalance in the data classes across clients decreases. 
%This allows us to test our method across a broad range of non-IID scenarios. 
We choose $\alpha = \{0.1, 0.5, 1.0\}$ in our experiments. 

\noindent\textbf{Model Heterogeneity}. Following FedTGP \cite{zhang2024fedtgp}, we simulate model heterogeneity by initializing a set of model architectures and assigning them to different clients. Specifically, we use “$\text{HtFE}_X$” to denote different settings, where “$\text{FE}_X$” refers to the number of heterogeneous feature extractors. Each client $i$ is assigned the ($i \mod X$)-th model architecture. We consider four model heterogeneity settings: $\text{HtFE}_2$, $\text{HtFE}_3$, $\text{HtFE}_4$, and $\text{HtFE}_9$. To ensure that feature outputs from all clients have the same dimensionality $d$, we append an average pooling layer after each feature extractor. $d$ is set to 512 by default.

\definecolor{c1}{RGB}{189,230,205}
\definecolor{c2}{RGB}{228,238,188}
\definecolor{c3}{RGB}{255,248,197}
\definecolor{c4}{RGB}{238,238,238}
\begin{table*}[tb]
	\setlength{\abovecaptionskip}{0.cm}
\setlength{\belowcaptionskip}{-0.cm}
\setlength\tabcolsep{2.0pt}
	\begin{center}
    	\renewcommand\arraystretch{0.8}
 % \scriptsize
 % \small
 \footnotesize
 \caption{Test accuracy (\%) of different methods under various non-IID scenarios on CIFAR-10, CIFAR-100, and Tiny ImageNet with $\text{HtFE}_9$. The top three results are highlighted as \colorbox{c1}{\textbf{first}}, \colorbox{c2}{second}, and \colorbox{c3}{third}, respectively.}
        \label{tab:effect of data hetero}
		\begin{tabular}{@{}cccccccccc@{}}
			\toprule
			&  \multicolumn{3}{c}{CIFAR-10}             &  \multicolumn{3}{c}{CIFAR-100}              & \multicolumn{3}{c}{Tiny ImageNet}           \\ \midrule
			Methods &  $\alpha=0.1$ & $\alpha=0.5$ & $\alpha=1.0$ &  $\alpha=0.1$ & $\alpha=0.5$ & $\alpha=1.0$ & $\alpha=0.1$ & $\alpha=0.5$ & $\alpha=1.0$ \\ \midrule
            \makecell{FedGen} &  \makecell{84.42\scriptsize$\pm$0.34} & \makecell{61.63\scriptsize$\pm$0.31} & \makecell{54.54\scriptsize$\pm$0.14} & \makecell{40.64\scriptsize$\pm$0.12} & \makecell{23.20\scriptsize$\pm$0.21} & \makecell{17.86\scriptsize$\pm$0.34} &  \makecell{30.86\scriptsize$\pm$0.22} & \makecell{14.57\scriptsize$\pm$0.06} & 
            \makecell{10.65\scriptsize$\pm$0.15} \\
            \makecell{FedGH} &  \makecell{83.59\scriptsize$\pm$0.16} & \makecell{60.87\scriptsize$\pm$0.26} & \makecell{54.83\scriptsize$\pm$0.06} & \makecell{40.42\scriptsize$\pm$0.17} & \makecell{22.68\scriptsize$\pm$0.21} & \makecell{17.64\scriptsize$\pm$0.14} &  \makecell{29.95\scriptsize$\pm$0.23} & \makecell{12.23\scriptsize$\pm$0.18} & 
            \makecell{\; 9.12\scriptsize$\pm$0.04} \\
            \makecell{LG-FedAvg} &  \makecell{84.53\scriptsize$\pm$0.13} & \makecell{61.31\scriptsize$\pm$0.09} & \makecell{55.26\scriptsize$\pm$0.11} & \makecell{42.56\scriptsize$\pm$0.23} & \makecell{25.03\scriptsize$\pm$0.16} & \makecell{19.25\scriptsize$\pm$0.24} &  \makecell{32.30\scriptsize$\pm$0.19} & \makecell{15.75\scriptsize$\pm$0.06} & 
            \makecell{11.79\scriptsize$\pm$0.05} \\
            \makecell{FML} &  \makecell{86.79\scriptsize$\pm$0.18} & \makecell{63.53\scriptsize$\pm$0.34} & \makecell{57.09\scriptsize$\pm$0.13} & \makecell{42.49\scriptsize$\pm$0.36} & \makecell{25.32\scriptsize$\pm$0.17} & \makecell{19.46\scriptsize$\pm$0.12} &  \makecell{32.43\scriptsize$\pm$0.07} & \makecell{\multicolumn{1}{>{\columncolor{c3}}c}{17.99\scriptsize$\pm$0.05}} & \makecell{\multicolumn{1}{>{\columncolor{c3}}c}{13.85\scriptsize$\pm$0.03}} \\
            \makecell{FedDistill} &  \makecell{85.93\scriptsize$\pm$0.11} & \makecell{62.35\scriptsize$\pm$0.15} & \makecell{55.75\scriptsize$\pm$0.14} & \makecell{42.47\scriptsize$\pm$0.16} & \makecell{\multicolumn{1}{>{\columncolor{c3}}c}{25.50\scriptsize$\pm$0.09}} & \makecell{\multicolumn{1}{>{\columncolor{c3}}c}{19.55\scriptsize$\pm$0.02}} &  \makecell{32.22\scriptsize$\pm$0.32} & \makecell{17.68\scriptsize$\pm$0.09} & \makecell{13.07\scriptsize$\pm$0.08} \\
            \makecell{FedKD} &  \makecell{\multicolumn{1}{>{\columncolor{c3}}c}{86.80\scriptsize$\pm$0.15}} & \makecell{63.86\scriptsize$\pm$0.16} & \makecell{57.05\scriptsize$\pm$0.40} & \makecell{42.54\scriptsize$\pm$0.49} & \makecell{24.78\scriptsize$\pm$0.10} & \makecell{19.20\scriptsize$\pm$0.16} &  \makecell{\multicolumn{1}{>{\columncolor{c3}}c}{32.79\scriptsize$\pm$0.08}} & \makecell{17.80\scriptsize$\pm$0.09} & \makecell{13.72\scriptsize$\pm$0.04} \\
            \makecell{FedMRL} &  \makecell{86.70\scriptsize$\pm$0.19} & \makecell{\multicolumn{1}{>{\columncolor{c3}}c}{64.06\scriptsize$\pm$0.14}} & \makecell{\multicolumn{1}{>{\columncolor{c3}}c}{57.11\scriptsize$\pm$0.30}} & \makecell{\multicolumn{1}{>{\columncolor{c3}}c}{42.82\scriptsize$\pm$0.37}} & \makecell{24.87\scriptsize$\pm$0.23} & \makecell{19.25\scriptsize$\pm$0.29} &  \makecell{31.43\scriptsize$\pm$0.22} & \makecell{16.22\scriptsize$\pm$0.14} & \makecell{12.48\scriptsize$\pm$0.19} \\
            \makecell{FedProto} &  \makecell{85.81\scriptsize$\pm$0.06} & \makecell{61.53\scriptsize$\pm$0.21} & \makecell{54.34\scriptsize$\pm$0.14} & \makecell{41.07\scriptsize$\pm$0.17} & \makecell{24.48\scriptsize$\pm$0.09} & \makecell{18.89\scriptsize$\pm$0.29} &  \makecell{31.52\scriptsize$\pm$0.20} & \makecell{16.96\scriptsize$\pm$0.10} & \makecell{12.51\scriptsize$\pm$0.05} \\
            \makecell{FedTGP} &  \makecell{85.73\scriptsize$\pm$0.03} & \makecell{61.60\scriptsize$\pm$0.31} & \makecell{53.96\scriptsize$\pm$0.25} & \makecell{41.37\scriptsize$\pm$0.01} & \makecell{24.43\scriptsize$\pm$0.17} & \makecell{18.33\scriptsize$\pm$0.13} &  \makecell{31.16\scriptsize$\pm$0.10} & \makecell{15.70\scriptsize$\pm$0.02} & \makecell{11.80\scriptsize$\pm$0.11} \\
            \makecell{AlignFed} &  \makecell{85.80\scriptsize$\pm$0.34} & \makecell{62.43\scriptsize$\pm$0.10} & \makecell{56.59\scriptsize$\pm$0.30} & \makecell{41.88\scriptsize$\pm$0.05} & \makecell{24.22\scriptsize$\pm$0.23} & \makecell{18.22\scriptsize$\pm$0.26} &  \makecell{30.77\scriptsize$\pm$0.13} & \makecell{14.54\scriptsize$\pm$0.15} & 
            \makecell{10.73\scriptsize$\pm$0.02} \\
            \midrule
            \makecell{\methodname-CLIP} &  \makecell{\multicolumn{1}{>{\columncolor{c2}}c}{87.34\scriptsize$\pm$0.08}} & \makecell{\multicolumn{1}{>{\columncolor{c2}}c}{64.76\scriptsize$\pm$0.21}} & \makecell{\multicolumn{1}{>{\columncolor{c1}}c}{\textbf{58.62\scriptsize$\pm$0.07}}} & \makecell{\multicolumn{1}{>{\columncolor{c2}}c}{45.61\scriptsize$\pm$0.19}} & \makecell{\multicolumn{1}{>{\columncolor{c1}}c}{\textbf{26.92\scriptsize$\pm$0.21}}} & \makecell{\multicolumn{1}{>{\columncolor{c1}}c}{\textbf{20.91\scriptsize$\pm$0.18}}} &  \makecell{\multicolumn{1}{>{\columncolor{c1}}c}{\textbf{34.82\scriptsize$\pm$0.13}}} & \makecell{\multicolumn{1}{>{\columncolor{c1}}c}{\textbf{18.33\scriptsize$\pm$0.13}}} & \makecell{\multicolumn{1}{>{\columncolor{c1}}c}{\textbf{14.43\scriptsize$\pm$0.07}}} \\
            \makecell{\methodname-BERT} &  \makecell{\multicolumn{1}{>{\columncolor{c1}}c}{\textbf{87.52\scriptsize$\pm$0.07}}} & \makecell{\multicolumn{1}{>{\columncolor{c1}}c}{\textbf{65.31\scriptsize$\pm$0.08}}} & \makecell{\multicolumn{1}{>{\columncolor{c2}}c}{58.42\scriptsize$\pm$0.12}} & \makecell{\multicolumn{1}{>{\columncolor{c1}}c}{\textbf{46.08\scriptsize$\pm$0.06}}} & \makecell{\multicolumn{1}{>{\columncolor{c2}}c}{26.66\scriptsize$\pm$0.11}} & \makecell{\multicolumn{1}{>{\columncolor{c2}}c}{20.53\scriptsize$\pm$0.15}} &  \makecell{\multicolumn{1}{>{\columncolor{c2}}c}{33.47\scriptsize$\pm$0.17}} & \makecell{\multicolumn{1}{>{\columncolor{c2}}c}{18.21\scriptsize$\pm$0.14}} & \makecell{\multicolumn{1}{>{\columncolor{c2}}c}{14.41\scriptsize$\pm$0.05}} \\
			
			\bottomrule
		\end{tabular}
        % \vspace{3pt}
	\end{center}
 \vspace{-15pt}
\end{table*}

\noindent\textbf{Implementation Details}. Our main experiments are conducted in the cross-silo scenario. The total number of clients $N = 20$, and all clients participate in training in each round. The batch size is set to 100 and the client local epoch $E_c=5$.
For each experiment, the number of communication rounds is set to 300, and the average test accuracy across all clients is calculated in each round. We report the highest average accuracy obtained over all rounds. 

For \methodname{}, we evaluate the performance of two different PLMs:
(1) CLIP: Since CLIP has already achieved image-text modality alignment during its pretraining phase, the text prototypes it generates are more readily beneficial to client-side image models.
(2) BERT: This represents a more general yet challenging setting, as BERT has not been exposed to image data during pretraining. Consequently, the modality gap between BERT and the client-side image model must be addressed.
In experiments, these two variants are referred to as \textbf{\methodname-CLIP} and \textbf{\methodname-BERT}, respectively.
More details are provided in the Appendix~C.
% Appendix~\ref{sec:detailed experiment setup}.
% \vspace{-6pt}

\subsection{Comparison with SOTA Methods}
\textbf{Performance under Different Data Heterogeneity Scenarios.}
%We first compare the performance of different methods across various non-IID settings. 
As shown in Table~\ref{tab:effect of data hetero}, the results reveal that:
(1) \methodname{} significantly outperforms existing prototype-based methods (FedProto, FedTGP, AlignFed) by up to 4.20\%, demonstrating that incorporating rich semantic relationships from the textual modality effectively enhances prototype quality, thereby improving client model performance.
(2)	\methodname{} consistently surpasses all SOTA methods, with a maximum performance gain of 3.26\%.
(3) The performance improvement of \methodname{} over SOTA methods is more pronounced under strong data heterogeneity, indicating that \methodname{} is more robust to heterogeneous data distributions. This robustness stems from the fact that prototypes constructed using LLMs and PLMs are less affected by local data distribution shifts. In addition, we also validate the performance of \methodname{} under different model heterogeneity scenarios. Please refer to Appendix~D for detail.
% Furthermore, these results suggest that under severe data heterogeneity, individual clients struggle to construct robust semantic relationships, making it increasingly crucial to acquire text-based semantic relationships from the server.

% \input{tables/3_experiment/SOTA_model}

% \noindent\textbf{Performance under Different Model Heterogeneity Scenarios.}
% %We further evaluate the effectiveness of our method under various model heterogeneity settings. 
% As presented in Table~\ref{tab:effect of model hetero}, our approach consistently outperforms all SOTA methods by up to 3.55\% across different model architectures, demonstrating its adaptability across diverse model structures.

\begin{figure*}[tb]
  \centering
  % 第一个图
  \begin{minipage}[t]{0.48\textwidth}
  	\setlength{\abovecaptionskip}{0.0cm}
\setlength{\belowcaptionskip}{-0.cm}
    \centering  % 图片全局居中
    % CIFAR-10 子图
    \begin{subfigure}[b]{0.48\linewidth}
        \centering
        \includegraphics[width=\linewidth]{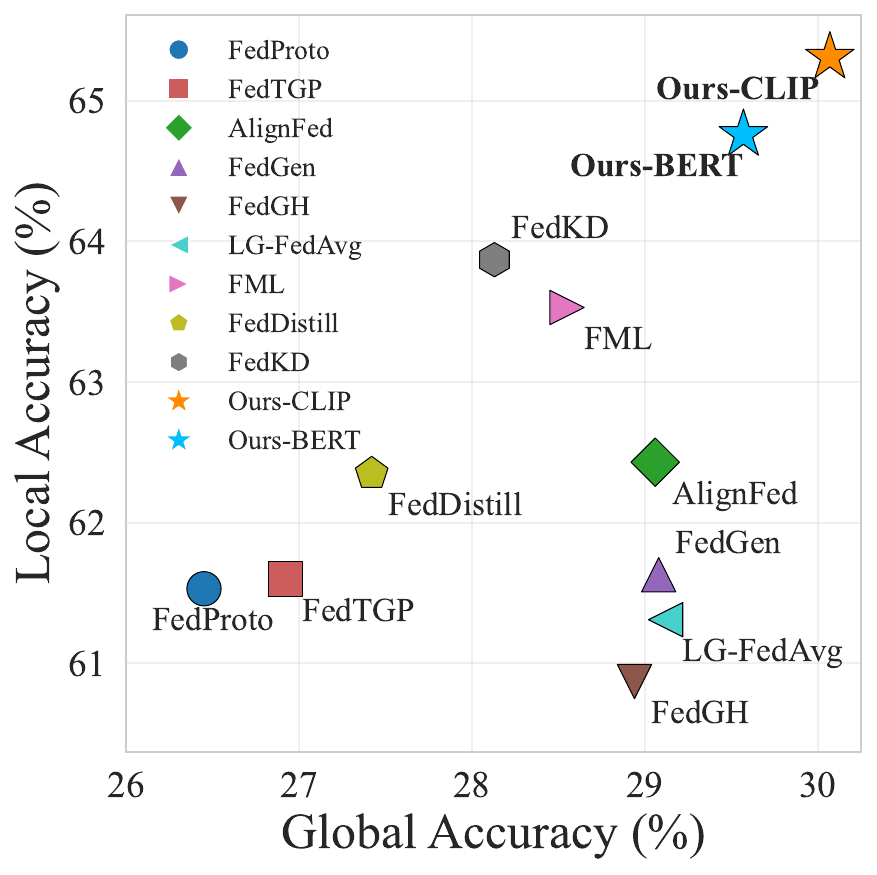}
        \caption{CIFAR-10}
    \end{subfigure}
    % CIFAR-100 子图
    \begin{subfigure}[b]{0.48\linewidth}
        \centering
\includegraphics[width=\linewidth]{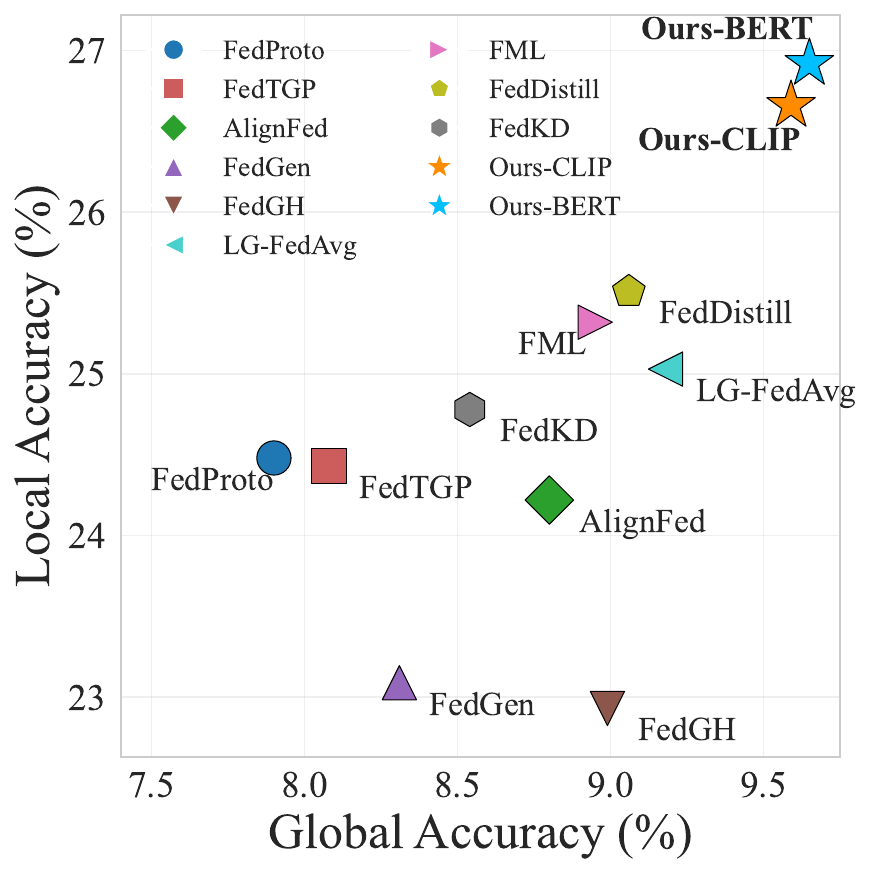}
        \caption{CIFAR-100}
    \end{subfigure}
    \caption{Comparison of local and global accuracies on CIFAR-10 and CIFAR-100 across different methods.}
    \label{fig:global acc}
  \end{minipage}
  \hfill % 填充水平间距
  % 第二个图
  \begin{minipage}[t]{0.48\textwidth}
         	\setlength{\abovecaptionskip}{0.0cm}
\setlength{\belowcaptionskip}{-0.cm} 
    \centering
\includegraphics[width=\linewidth]{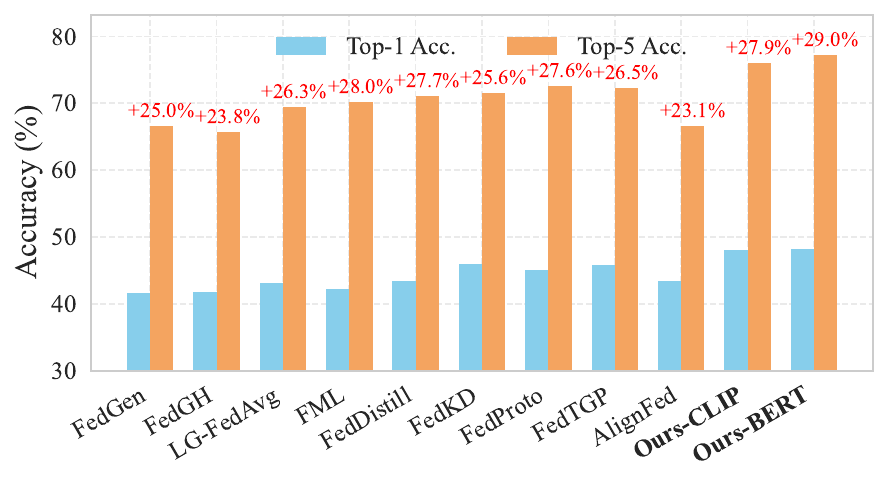}
    \caption{Comparison of Top-1 and Top-5 accuracies on CIFAR-100 across different methods.}
    \label{fig:top5 acc}
  \end{minipage}
  \vspace{-10pt}
\end{figure*}

\noindent\textbf{Performance on General Test Set.}
% 在Cifar10 和Cifar 100， dir 0.5，HtFE3上画一个二维坐标图。横轴是global accuracy，纵轴是local accuracy。
Retaining model generalization is a critical concern in FL. Current HtFL methods mostly focus on boosting personalization (\ie, local accuracy) while overlooking generalization on balanced datasets (\ie, global accuracy).
We plot both local and global accuracy for different methods in Fig.~\ref{fig:global acc}. As shown, some approaches (\eg, FedKD and FML) improve local accuracy but sacrifice global accuracy. By contrast, our method achieves both strong local and global accuracy.

\noindent\textbf{Top-5 Accuracy of Different Methods.}
% 在Cifar 100， dir 0.5，HtFE3上画一个柱形图。分别展示top 1 acc和top 5 acc的值。说明我们的方法提升top 5 acc更显著。
%Top-5 accuracy is a commonly used metric in multi-class classification tasks, indicating the fraction of samples for which the correct label appears among the top five predictions. 
In \methodname{}, the introduction of inter-class semantic relationships from the text modality helps clients learn class similarities more effectively. Even if a sample is misclassified, it is more likely to be placed in a closely related class.
We compare Top-5 accuracy on CIFAR-100 in Fig.~\ref{fig:top5 acc}. The results show that \methodname{} achieves a more significant improvement in top-5 accuracy compared to SOTA methods. This highlights the effectiveness and necessity of incorporating textual semantic relationships in FL.
% \begin{figure}[htp]
% 	\setlength{\abovecaptionskip}{0.cm}
% \setlength{\belowcaptionskip}{-0.cm}
%     \centering  % 图片全局居中
%     % CIFAR-10 子图
%     \begin{subfigure}[b]{0.485\linewidth}
%         \centering
%         \includegraphics[width=\linewidth]{pictures/Cifar10_dir0.5_HtFE9_global_accuracy.pdf}
%         \caption{CIFAR-10}
%     \end{subfigure}
%     % CIFAR-100 子图
%     \begin{subfigure}[b]{0.48\linewidth}
%         \centering
%         \includegraphics[width=\linewidth]{pictures/Cifar100_dir0.5_HtFE9_global_accuracy.pdf}
%         \caption{CIFAR-100}
%     \end{subfigure}
%     \caption{Comparison of local and global accuracy on CIFAR-10 and CIFAR-100 under the Dir(0.5) settings using $\text{HtFE}_9$.}
%     \label{fig:global acc}
% \end{figure}

\subsection{Ablation Study}\label{sec:ablation}
\textbf{Effect of Different Modules.}
%We evaluate the impact of different modules in \methodname{}. 
The results are presented in Table~\ref{tab:ablation}. Here, -BERT and -CLIP represent \methodname-BERT and \methodname-CLIP, respectively.
\textbf{ID 1:} Local training on each client (baseline).
\textbf{ID 2:} Incorporating handcrafted prompts to generate prototypes. The results show that even basic textual semantic information significantly boosts performance.
\textbf{ID 3:} Using an LLM to generate finer-grained class descriptions as prompts. By enriching class descriptions, the text prototype gains more semantic information, leading to better inter-class semantic relationships and improved guidance for client-side image classification.
\textbf{ID 4:} Adding trainable prompts (on top of ID 2) to align the text prototype with the client’s image prototype. This yields a significant improvement (+1.9\%) for -BERT, as BERT has not seen image data during pretraining, creating a modality gap. By adding trainable prompts, the alignment between text and image modalities is improved. In contrast, CLIP, which already aligns the modalities during pretraining, shows less improvement from adding trainable prompts.
\textbf{ID 5:} Building on ID 4 by incorporating LLM-generated prompts. This further enhances performance with both text encoders, highlighting the importance of introducing richer semantic information through LLMs.
% \begin{wraptable}{r}{0.48\textwidth}
\begin{table}[tb]
\setlength{\abovecaptionskip}{0.cm}
\setlength{\belowcaptionskip}{-0.cm}
     \renewcommand\arraystretch{0.6}

\setlength\tabcolsep{4pt}
% \vspace{-4pt}

 \small
 \caption{Ablation study on CIFAR-100 illustrating the effectiveness of different modules.}
\label{tab:ablation}
	\begin{center}
				\begin{tabular}{lccccc}
\toprule
ID & Proto. & LLM & Trainable & -BERT & -CLIP   \\ \midrule
1 & & & & 23.47$\pm$\scriptsize 0.13 & 23.47$\pm$\scriptsize 0.13  \\
2 & \checkmark & &  \multicolumn{1}{c}{} & 24.63$\pm$\scriptsize0.22 & 26.41$\pm$\scriptsize 0.11 \\
3 & \checkmark & \checkmark &  \multicolumn{1}{c}{} & 25.76$\pm$\scriptsize 0.18 & 26.99$\pm$\scriptsize 0.10 \\
% \midrule
4 & \checkmark &  & \checkmark &  26.53$\pm$\scriptsize 0.18 & 26.54$\pm$\scriptsize 0.14 \\
5 & \checkmark & \checkmark & \multicolumn{1}{c}{\checkmark} & 27.01$\pm$\scriptsize 0.12 & 27.34$\pm$\scriptsize 0.11 \\

\bottomrule
\end{tabular}
	\end{center}

\vspace{-15pt}
\end{table}
% \end{wraptable}

\noindent\textbf{The Impact of Semantic Information Richness.}
%In this experiment, we investigate how the level of semantic richness in prompts affects model performance. 
We compare three types of prompts: \textit{Manual}: A simple template, “A photo of a \{CLASS\}.”
\textit{Short}: A brief description for each class generated by an LLM.
\textit{Long}: A detailed description for each class generated by an LLM.

\begin{figure}[tb]
\setlength{\abovecaptionskip}{0.0cm}
\setlength{\belowcaptionskip}{-0.cm}
    \centering
% \vspace{-1pt}
    \begin{adjustbox}{valign=b}
        \begin{minipage}[t]{0.47\linewidth}
  \setlength{\abovecaptionskip}{0.1cm}
\setlength{\belowcaptionskip}{-0.cm}
    % \centering  % 图片全局居中
    % CIFAR-10 子图
    % \begin{subfigure}[b]{0.490\linewidth}
       \centering
\includegraphics[width=\linewidth]{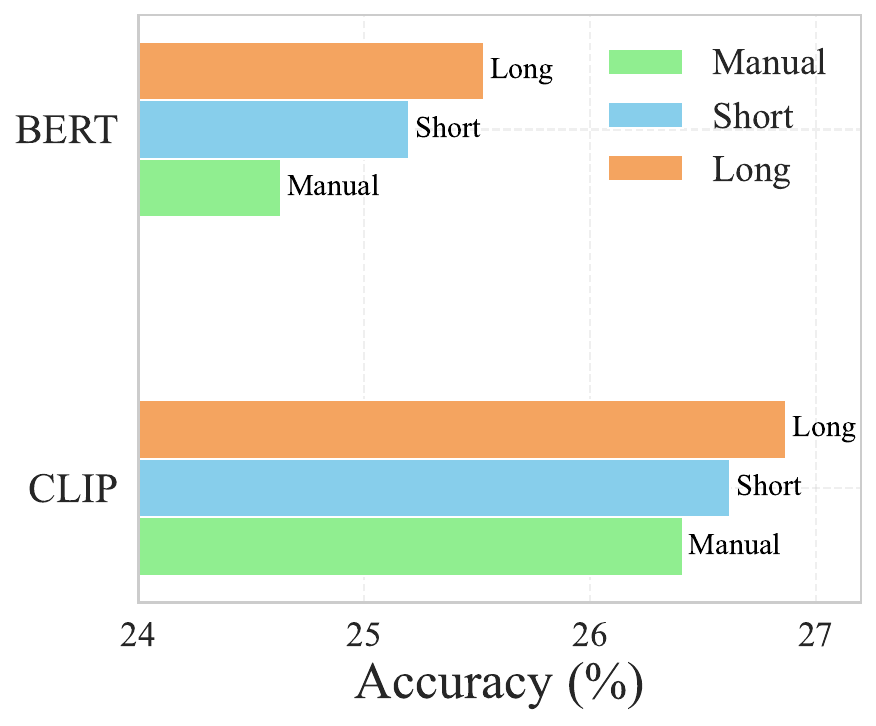}
        % \caption{Effect of prompt type for text encoder.}
        % \label{fig:prompt type}
        % \caption{Effect of prompt type for text encoder.}
        % \label{fig:prompt type}
  \end{minipage}
    \end{adjustbox}
    \hfill
    \begin{adjustbox}{valign=b}
        \begin{minipage}[t]{0.46\linewidth}
            \scriptsize
            \setlength\tabcolsep{1pt}
            \begin{tabular}{lccc}
					\toprule
                      CIFAR-10 & Dir(0.1) & Dir(0.5) & Dir(1.0) \\
                    \midrule
                    GPT-4o & 87.34 & 64.76 & 58.62 \\
                    Claude & 86.80 & 65.36 & 58.23 \\
                    Gemini & 86.92 & 65.00 & 58.69 \\
                    LLaMa3 & 86.89 & 64.97 & 58.42 \\
                    \midrule
                    CIFAR-100 & Dir(0.1) & Dir(0.5) & Dir(1.0) \\
                    \midrule
                    GPT-4o & 45.61 & 26.92 & 20.91 \\
                    Claude & 45.80 & 26.97 & 20.78 \\
                    Gemini & 45.66 & 27.24 & 20.58 \\
                    LLaMa3 & 45.61 & 26.41 & 21.33 \\
                    
                    \bottomrule
				\end{tabular}
                % \caption{Effect of LLM Choice on Overall Performance.}
                % \label{tab:LLM type}
        \end{minipage}
    \end{adjustbox}
    \caption{\textbf{Left:} Effect of prompt type on the text encoder. \textbf{Right:} Comparison of different LLMs for prompt generation.}
    \label{fig:prompt and llm type}
    \vspace{-15pt}
\end{figure}

As shown in the Fig.~\ref{fig:prompt and llm type} (left), adding richer semantic information to the prompts consistently improves model performance. This finding underscores both the effectiveness of textual semantic cues for client-side classification and the importance of leveraging LLMs to enrich semantic content.

\noindent\textbf{Robustness to Different LLM Choices.} 
To further examine whether FedTSP depends on a specific LLM, we compare several representative LLMs for generating textual descriptions used in prompt construction. Specifically, we evaluate three commercial models, GPT-4o (used in our main experiments), Claude-4-sonnet, and Gemini-2.5-pro, as well as an open-source model, LLaMA3-8B-Instruct. As shown in Fig.~\ref{fig:prompt and llm type} (right), FedTSP exhibits highly consistent performance across all four models under different settings. These results demonstrate that FedTSP does not rely on the strength or scale of a particular LLM. We attribute this robustness to two factors. First, modern LLMs share overlapping world knowledge about common visual categories, which enables them to produce similarly informative descriptions. Second, FedTSP employs multiple ($k = 3$) textual descriptions for each class, covering diverse semantic aspects such as appearance, functionality, and environment.

%cvpr

\subsection{Convergence Speed}
%In this section, we empirically verify the convergence speed of FedTSP. 

\begin{figure}[ht]
\centering
  \begin{minipage}[t]{\linewidth}
   \setlength{\abovecaptionskip}{0.0cm}
\setlength{\belowcaptionskip}{-0.cm}
    \centering  % 图片全局居中

    % CIFAR-10 子图
    \begin{subfigure}[b]{0.49\linewidth}
        \centering
        \includegraphics[width=\linewidth,height=0.71\linewidth]{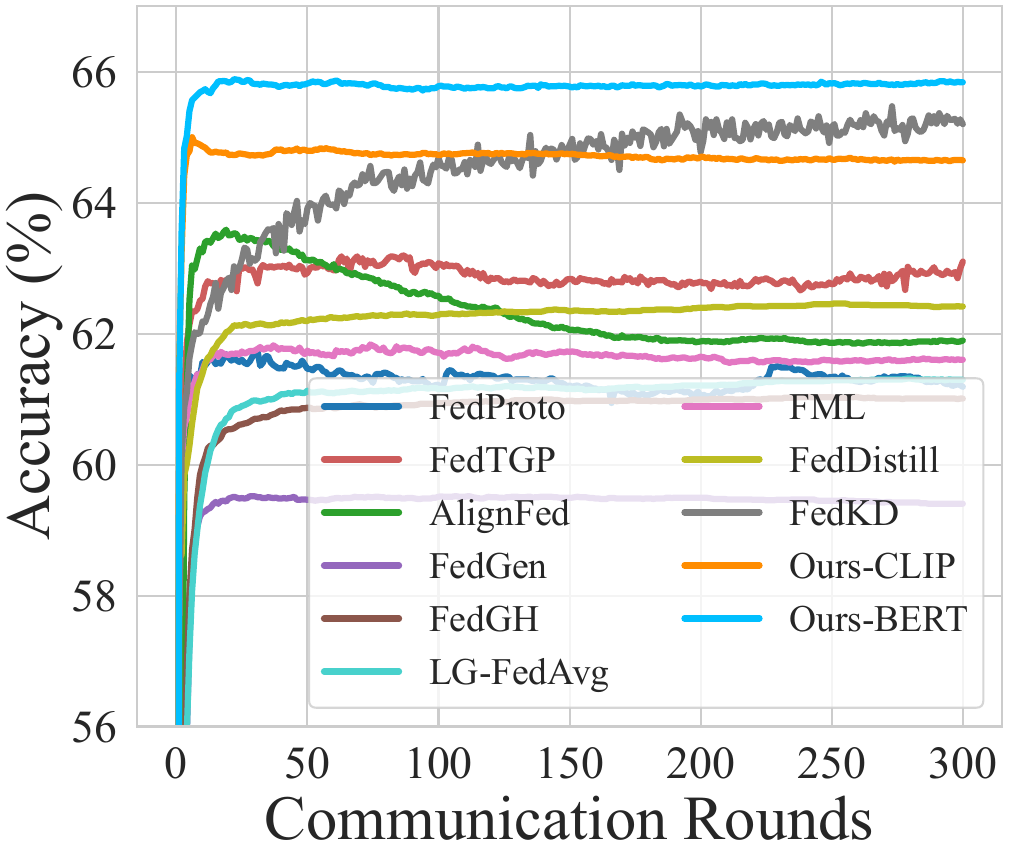}
        \caption{CIFAR-10}
    \end{subfigure}
    % CIFAR-100 子图
    \begin{subfigure}[b]{0.49\linewidth}
        \centering
        \includegraphics[width=\linewidth,height=0.72\linewidth]{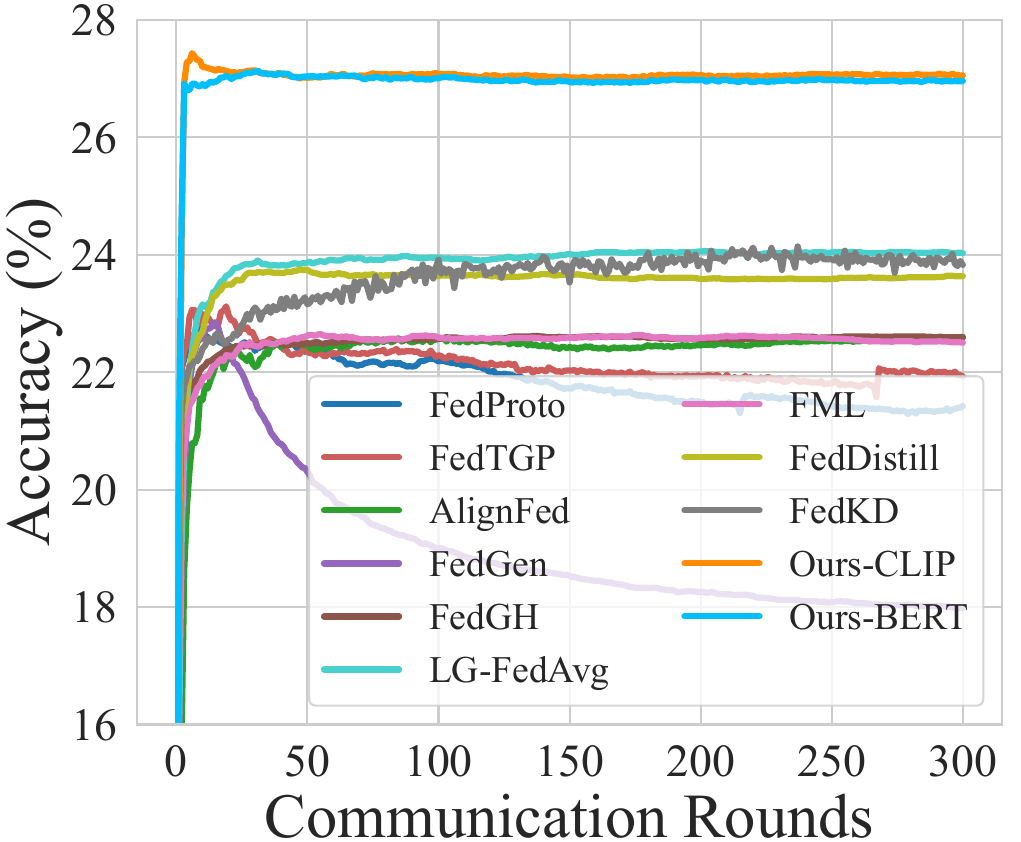}
        \caption{CIFAR-100}
    \end{subfigure}
    
    \caption{The test accuracy (\%) curves of different methods on CIFAR-10 and CIFAR-100.}
    \label{fig:convengence curve}
  \end{minipage}
  \vspace{-15pt}
\end{figure}

As shown in Fig.~\ref{fig:convengence curve}, FedTSP converges significantly faster than SOTA methods. This superiority arises from PLM, which provides a high-quality global prototype at the outset and thus spares clients from having to perform a mutual trade-off to obtain a global prototype (\eg, FedProto and FedTGP).
Beyond its faster convergence, FedTSP also exhibits highly stable learning curves, in contrast to certain methods that suffer large training fluctuations due to inter-client interference (\eg, FedKD and FedProto) or exhibit reduced test accuracy from overfitting local client data (\eg, AlignFed and FedGen). 
Overall, FedTSP not only effectively boosts accuracy but also significantly accelerates convergence, thereby reducing communication overhead.
% \begin{figure}[htb]
% 	\setlength{\abovecaptionskip}{0.cm}
% \setlength{\belowcaptionskip}{-0.cm}
%     \centering  % 图片全局居中

%     % CIFAR-10 子图
%     \begin{subfigure}[b]{0.48\linewidth}
%         \centering
%         \includegraphics[width=\linewidth]{pictures/Cifar10_dir_0.5_balance_20_HtFE3_gr300_ep5_bs100_nc20_seed0.pdf}
%         \caption{CIFAR-10}
%     \end{subfigure}
%     % CIFAR-100 子图
%     \begin{subfigure}[b]{0.485\linewidth}
%         \centering
%         \includegraphics[width=\linewidth]{pictures/Cifar100_dir_0.5_balance_20_HtFE3_gr300_ep5_bs100_nc20_seed0.pdf}
%         \caption{CIFAR-100}
%     \end{subfigure}
    
%     \caption{The test accuracy (\%) curve of different methods on CIFAR-10 and CIFAR-100 under Dir(0.5) using $\text{HtFE}_3$.}
%     \label{fig:convengence curve}
% \end{figure}

% cvpr25

\subsection{Results in Model-Homogeneous Settings}\label{sec:GFL and PFL expe}
We compare \methodname{} with SOTA \FPL{} methods from GFL and PFL in model-homogeneous scenarios. The experimental results on CIFAR-100 are presented in Table~\ref{tab:GFL and PFL}, indicating that \methodname{} outperforms most SOTA methods in both settings. Note that -FT refers to fine-tuning the classifier of the global model, as we observe that fine-tuning the classifier always outperforms full fine-tuning.

% \begin{wraptable}{r}{0.48\textwidth}
\begin{table}[htb]
% \vspace{-170pt}
% \vspace{-9pt}
	\setlength{\abovecaptionskip}{0.cm}
\setlength{\belowcaptionskip}{-0.cm}
    \setlength\tabcolsep{2pt}
         \renewcommand\arraystretch{0.4}
 \small
	\begin{center}
    \caption{Test accuracy (\%) of different methods in GFL and PFL settings under Dirichlet non-IID partition on CIFAR-100. }
 \label{tab:GFL and PFL}
		\begin{tabular}{cccc}
			\toprule
			Settings & $\alpha=0.1$ & $\alpha=0.5$ & $\alpha=1.0$ \\
		\midrule
            \multicolumn{4}{c}{General Federated Learning} \\
            \midrule
            \makecell{FedAvg} & \makecell{34.47\scriptsize$\pm$0.85} & \makecell{31.96\scriptsize$\pm$0.70} & \makecell{33.20\scriptsize$\pm$0.62}  \\
            \makecell{FedBABU} & \makecell{\multicolumn{1}{>{\columncolor{c3}}c}{36.37\scriptsize$\pm$0.76}} & \makecell{33.02\scriptsize$\pm$1.15} & \makecell{33.61\scriptsize$\pm$0.35}  \\
            \makecell{FedETF} & \makecell{33.84\scriptsize$\pm$1.05} & \makecell{31.10\scriptsize$\pm$1.05} & \makecell{32.24\scriptsize$\pm$0.24}  \\
            \makecell{FedFA} & \makecell{35.16\scriptsize$\pm$1.08} & \makecell{\multicolumn{1}{>{\columncolor{c3}}c}{34.07\scriptsize$\pm$0.78}} & \makecell{\multicolumn{1}{>{\columncolor{c1}}c}{\textbf{34.69\scriptsize$\pm$1.58}}}  \\
            \textbf{Ours-CLIP} & \makecell{\multicolumn{1}{>{\columncolor{c1}}c}{\textbf{37.98\scriptsize$\pm$1.65}}} & \makecell{\multicolumn{1}{>{\columncolor{c2}}c}{34.10\scriptsize$\pm$1.01}} & \makecell{\multicolumn{1}{>{\columncolor{c2}}c}{34.38\scriptsize$\pm$0.33}}  \\
			\textbf{Ours-BERT} &\makecell{\multicolumn{1}{>{\columncolor{c2}}c}{37.27\scriptsize$\pm$1.54}} & \makecell{\multicolumn{1}{>{\columncolor{c1}}c}{\textbf{34.58\scriptsize$\pm$0.56}}} & \makecell{\multicolumn{1}{>{\columncolor{c3}}c}{34.30\scriptsize$\pm$0.51}}  \\
			\midrule
                \multicolumn{4}{c}{Personalized Federated Learning} \\
                \midrule
                % Local & \makecell{\scriptsize$\pm$} & \makecell{\scriptsize$\pm$} & \makecell{\scriptsize$\pm$}  \\
			% \makecell{FedCAC} & \makecell{57.22\scriptsize$\pm$1.52} & \makecell{38.64\scriptsize$\pm$0.63} & \makecell{32.59\scriptsize$\pm$0.32}  \\
            \makecell{FedAvg-FT} & \makecell{60.08\scriptsize$\pm$0.90} & \makecell{40.72\scriptsize$\pm$0.55} & \makecell{37.54\scriptsize$\pm$0.46}  \\
            \makecell{FedBABU-FT} & \makecell{\multicolumn{1}{>{\columncolor{c3}}c}{60.10\scriptsize$\pm$0.54}} & \makecell{\multicolumn{1}{>{\columncolor{c3}}c}{40.95\scriptsize$\pm$0.92}} & \makecell{\multicolumn{1}{>{\columncolor{c3}}c}{38.05\scriptsize$\pm$0.89}}  \\
            \makecell{FedETF-Per} & \makecell{57.00\scriptsize$\pm$0.92} & \makecell{36.87\scriptsize$\pm$1.97} & \makecell{35.18\scriptsize$\pm$1.32}  \\
            \makecell{FedFA-FT} & \makecell{42.62\scriptsize$\pm$0.59} & \makecell{33.39\scriptsize$\pm$0.41} & \makecell{33.88\scriptsize$\pm$0.79}  \\
            \makecell{FedPCL} & \makecell{42.61\scriptsize$\pm$0.60} & \makecell{18.19\scriptsize$\pm$0.61} & \makecell{12.68\scriptsize$\pm$0.14}  \\
            \makecell{FedPAC} & \makecell{58.19\scriptsize$\pm$0.21} & \makecell{38.20\scriptsize$\pm$0.13} & \makecell{34.44\scriptsize$\pm$0.72}  \\
               \textbf{Ours-CLIP-FT} & \makecell{\multicolumn{1}{>{\columncolor{c1}}c}{\textbf{62.30\scriptsize$\pm$1.14}}} & \makecell{\multicolumn{1}{>{\columncolor{c1}}c}{\textbf{42.05\scriptsize$\pm$0.85}}} & \makecell{\multicolumn{1}{>{\columncolor{c2}}c}{38.30\scriptsize$\pm$0.37}}  \\
			\textbf{Ours-BERT-FT} & \makecell{\multicolumn{1}{>{\columncolor{c2}}c}{61.05\scriptsize$\pm$0.40}} & \makecell{\multicolumn{1}{>{\columncolor{c2}}c}{41.40\scriptsize$\pm$0.43}} & \makecell{\multicolumn{1}{>{\columncolor{c1}}c}{\textbf{38.50\scriptsize$\pm$0.65}}}  \\
			
			\bottomrule
		\end{tabular}
        % \vspace{3pt}
	\end{center}
    \vspace{-10pt}
% \end{wraptable}
\end{table}

\subsection{Privacy-Preserving Evaluation}\label{sec:privacy expe}
In this section, we evaluate whether the privacy-preserving extension described in Section~\ref{sec:privacy protection} affects model performance.
We set $C=2.5$ and $\delta=10^{-5}$, and vary $\epsilon \in [1, 2, 4, 8]$, where a small $\epsilon$ implies stronger privacy. As shown in Table~\ref{tab:privacy}, FedTSP remains robust under all settings. Even with strong privacy constraints ($\epsilon=1$), the accuracy degradation is negligible compared with the non-private baseline. These results confirm that the proposed local DP mechanism effectively protects class information while preserving the semantic alignment and overall learning performance of FedTSP.
\begin{table}[]
	\setlength{\abovecaptionskip}{0.cm}
\setlength{\belowcaptionskip}{-0.cm}
     \renewcommand\arraystretch{0.6}
% \vspace{7pt}
 % \scriptsize
 \footnotesize
	\begin{center}
    \caption{Accuracy (\%) of \methodname{} under different privacy levels on CIFAR-100.}
 \label{tab:privacy}
		\begin{tabular}{cccc}
			\toprule
			 & Dir(0.1) & Dir(0.5) & Dir(1.0) \\
            \midrule
            \makecell{\methodname{} (origin)} & \makecell{45.61\scriptsize$\pm$0.19} & \makecell{26.92\scriptsize$\pm$0.21} & \makecell{20.91\scriptsize$\pm$0.18}  \\
            \makecell{\methodname{} + DP ($\epsilon=8$)} & \makecell{45.40\scriptsize$\pm$0.14} & \makecell{26.90\scriptsize$\pm$0.18} & \makecell{20.68\scriptsize$\pm$0.22}  \\
            \makecell{\methodname{} + DP ($\epsilon=4$)} & \makecell{45.36\scriptsize$\pm$0.22} & \makecell{26.75\scriptsize$\pm$0.11} & \makecell{20.83\scriptsize$\pm$0.16}  \\
            \makecell{\methodname{} + DP ($\epsilon=2$)} & \makecell{45.46\scriptsize$\pm$0.31} & \makecell{26.40\scriptsize$\pm$0.22} & \makecell{20.93\scriptsize$\pm$0.23}  \\
            \makecell{\methodname{} + DP ($\epsilon=1$)} & \makecell{45.28\scriptsize$\pm$0.29} & \makecell{26.34\scriptsize$\pm$0.24} & \makecell{20.53\scriptsize$\pm$0.16} \\
			
			\bottomrule
		\end{tabular}
        % \vspace{3pt}
	\end{center}
    \vspace{-15pt}
\end{table}

\section{Conclusion}
This work revisits a long-overlooked challenge in \FPL{}: clients trained on scarce and highly non-IID data struggle to recover reliable inter-class semantic relationships, and existing prototype-based methods often unintentionally disrupt such structure. To address this limitation, we introduce \methodname{}, the first framework to inject global semantic priors from textual modality into FL. By generating fine-grained class descriptions with an LLM and deriving textual embeddings from a PLM, \methodname{} constructs prototypes that preserve rich semantic topology. In addition, its trainable prompts offer an effective mechanism for cross-modal alignment. Extensive experiments across HtFL, GFL, and PFL demonstrate that \methodname{} consistently enhances both global and local performance and accelerates convergence.

% NeurIPS 2025
% We identify that existing \FPL{} methods primarily focus on increasing the discrimination of prototypes, often at the cost of disrupting inter-class semantic relationships. To address this issue, we propose \methodname{}, which leverages an LLM on the server to generate fine-grained descriptions for each class and employs a PLM to construct semantically rich textual prototypes.
% Furthermore, \methodname{} introduces trainable prompts to bridge the gap between the PLM and client image models, ensuring that the generated prototypes better guide local client tasks. By aligning their local features with these prototypes, clients can effectively learn inter-class semantic relationships from the textual modality.
% Extensive experiments demonstrate that \methodname{} consistently outperforms SOTA methods across HtFL, GFL, and PFL settings.
% We elaborate on the limitations and outline future work in Appendix~F.
% Appendix~\ref{sec:limitation}

\section*{Acknowledgments}
This work was supported by the National Key R\&D Program of China under Grant 2023YFB4503700, the National Natural Science Foundation of China under Grants 62372028 and 62372027, and the Fundamental Research Funds for the Central Universities.
{
    \small
    \bibliographystyle{ieeenat_fullname}
    \bibliography{main}
}

% WARNING: do not forget to delete the supplementary pages from your submission 
% \input{sec/X_suppl}

\end{document}